\documentclass[11pt]{article}
\usepackage{aclbib}
\usepackage{amsmath, amsfonts}
\usepackage{graphicx}
\usepackage{setspace}
\usepackage{tabularx}

\usepackage{url}

\usepackage[margin=1.5in]{geometry}

\newcommand{\abs}[1]{\lvert#1\rvert}
\newcommand{\norm}[1]{\lVert#1\rVert}

\title{Robust Logistic Regression using Shift Parameters}

\author{Julie Tibshirani \\
  Stanford University \\
  Stanford, CA 94305, USA \\
  {\tt jtibs@cs.stanford.edu} \\ \\
  Master's Report, Stanford CS Department \\
  Advisor: Christopher Manning}
\date{}

\begin{document}
\maketitle

\begin{abstract}
Annotation errors can significantly hurt classifier performance, yet datasets are only growing noisier with the increased use of Amazon Mechanical Turk and techniques like distant supervision that automatically generate labels. In this paper, we present a robust extension of logistic regression that incorporates the possibility of mislabelling directly into the objective. Our model can be trained through nearly the same means as logistic regression, and retains its efficiency on high-dimensional datasets. Through named entity recognition experiments, we demonstrate that our approach can provide a significant improvement over the standard model when annotation errors are present.
\end{abstract}

\section{Introduction}
Almost any large dataset has annotation errors, especially those complex, nuanced datasets commonly used in natural language processing. Low-quality annotations have become even more common in recent years with the rise of Amazon Mechanical Turk, as well as methods like distant supervision and co-training that involve automatically generating training data.

Although small amounts of noise may not be detrimental, in some applications the level can be high: upon manually inspecting a relation extraction corpus commonly used in distant supervision,~\newcite{Riedel:10} report a 31\% false positive rate. In cases like these, annotation errors have frequently been observed to hurt performance. \newcite{Dingare:05}, for example, conduct error analysis on a system to extract relations from biomedical text, and observe that over half of the system's errors could be attributed to inconsistencies in how the data was annotated. Similarly, in a case study on co-training for natural language tasks, ~\newcite{Pierce:01} find that the degradation in data quality from automatic labelling prevents these systems from performing comparably to their fully-supervised counterparts.

Despite this prevalence, little work has been done in designing models that are aware of annotation errors. Moreover much of the previous work focuses on heuristic techniques to filter the data before training, which might discard valuable examples simply because they do not fit closely with the model assumptions.

In this work we argue that incorrect examples should be explicitly modelled during training, and present a simple extension of logistic regression that incorporates the possibility of mislabelling directly into the objective. Our model introduces sparse `shift parameters' to allow datapoints to slide along the sigmoid, changing class if appropriate. It has a convex objective, can handle high-dimensional data, and we show it can be efficiently trained with minimal changes to the logistic regression pipeline.

Experiments on large, noisy NER datasets show that our method can provide an improvement over standard logistic regression, both in manually and automatically annotated settings. The model also provides a means to identify which examples were mislabeled: through experiments on biological data, we demonstrate how our method can be used to accurately identify annotation errors. This robust extension of logistic regression shows promise in handling incorrect labels, while remaining efficient on large, high-dimensional datasets.

\section{Related Work}
Much of the previous work on dealing with annotation errors centers around filtering the data before training.  ~\newcite{Brodley:99} introduce what is perhaps the simplest form of supervised filtering: they train various classifiers, then record their predictions on a different part of the train set and eliminate contentious examples. In a similar vein, ~\newcite{Venkataraman:04} filter using SVMs, training on different subsets of the feature space to create multiple `views' of the data. 

\begin{figure}
\centering
\hspace{-0.8cm}
\includegraphics[width=13cm]{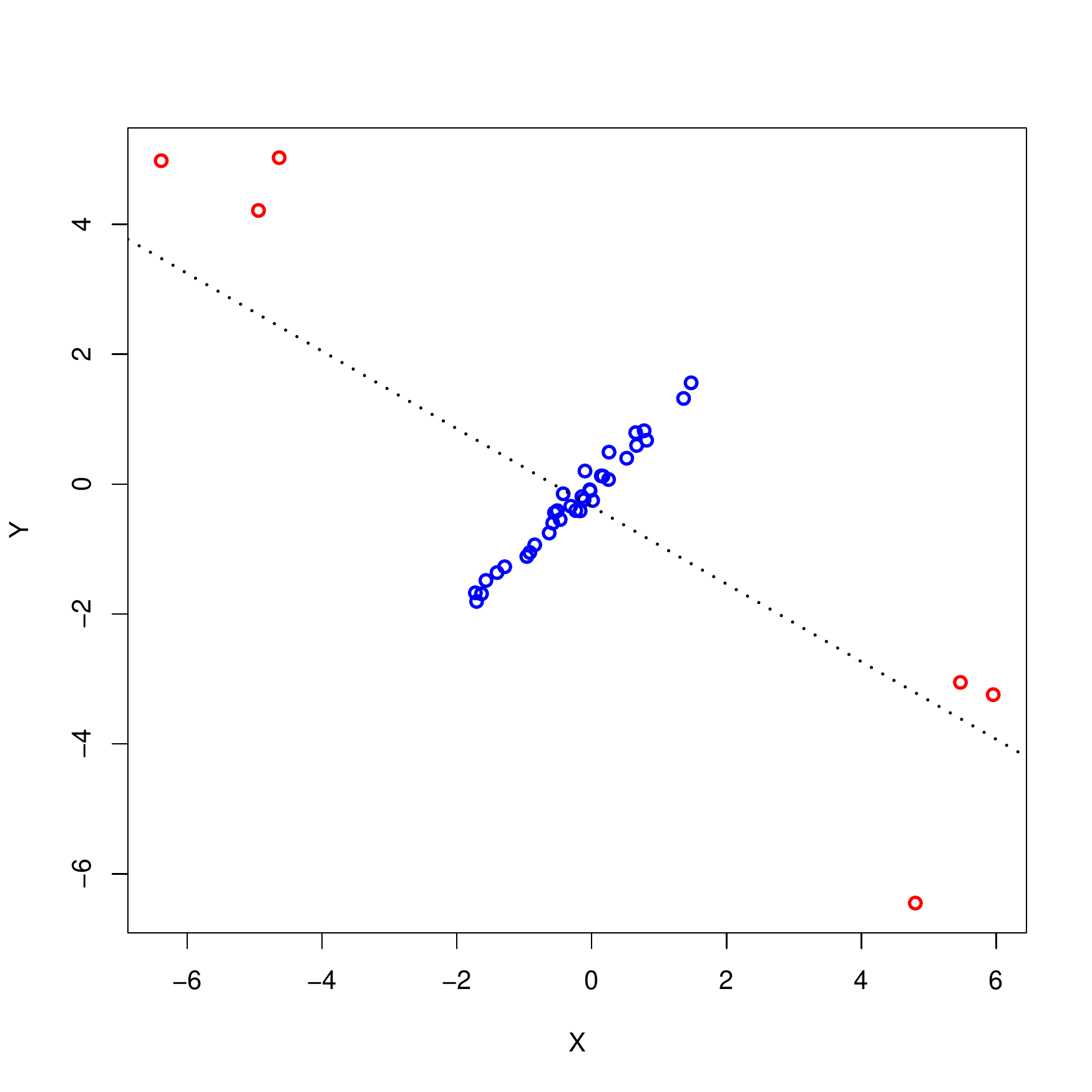}
\vspace{-0.5cm}

\caption{Fit resulting from running linear regression on the given data, which includes both clean examples (blue), and outliers (red). The outliers have such a large influence on the fit that they \emph{mask} each other's presence, and no longer appear that unusual. They also \emph{swamp} the procedure, so that some clean examples now begin to look suspicious.}
\end{figure}

One obvious issue with these methods is that the noise-detecting classifiers are themselves trained on noisy labels. Such methods may suffer from well-known effects like \emph{masking}, where several mislabelled examples `mask' each other and go undetected, and \emph{swamping}, in which the mislabelled points are so influential that they cast doubt on the correct examples \cite{She:11}. Figure 1 gives an example of these phenomena in the context of linear regression. Unsupervised filtering tries to avoid this problem by clustering training instances based solely on their features, then using the clusters to detect labelling anomalies \cite{Rebbapragada:09}. Recently, ~\newcite{Intxaurrondo:13} applied this approach to distantly-supervised relation extraction, using heuristics such as the number of mentions per tuple to eliminate suspicious examples.

Unsupervised filtering, however, relies on the perhaps unwarranted assumption that examples with the same label lie close together in feature space. Moreover filtering techniques in general may not be well-justified: if a training example does not fit closely with the current model, it is not necessarily mislabeled. Although the instance itself has a low-likelihood, it might represent an important exception that would improve the overall fit. An example may also appear unusual simply because we have made poor modelling assumptions, and in discarding it valuable information could be lost.

Perhaps the most promising approaches are those that directly model annotation errors, handling mislabelled examples as they train. This way, there is an active trade-off between fitting the model and identifying suspected errors. ~\newcite{Bootkrajang:12} present an extension of logistic regression that models annotation errors through flipping probabilities. For each example the authors posit a hidden variable representing its true label, and assume this label has a probability of being flipped before it is observed. While intuitive, this approach has shortcomings of its own: the objective function is nonconvex and the authors note that local optima are an issue, and the model can be difficult to fit when there are many more features than training examples.

The field of `robust statistics' seeks to develop estimators that are not unduly effected by deviations from the model assumptions~\cite{Huber:09}. Since mislabelled points are one type of outlier, this goal is naturally related to our interest in dealing with noisy data, and it seems many of the existing techniques would be relevant. A common strategy is to use a modified loss function that gives less influence to points far from the boundary, and several models along these lines have been proposed ~\cite{Ding:10,Masnadi-Shirazi:10}. Unfortunately these approaches require optimizing nonstandard, often nonconvex objectives, and fail to give insight into which datapoints are mislabeled.

In a recent advance,~\newcite{She:11} demonstrate that introducing a regularized `shift parameter' per datapoint can help increase the robustness of linear regression. ~\newcite{Candes:09} propose a similar approach for principal component analysis, while ~\newcite{Wright:09} explore its effectiveness in sparse signal recovery. In this work we adapt the technique to logistic regression. To the best of our knowledge, we are the first to experiment with adding `shift parameters' to logistic regression and demonstrate that the model is especially well-suited to the type of high-dimensional, noisy datasets commonly used in NLP.

There is a growing body of literature on learning from several annotators, each of whom may be inaccurate ~\cite{Bachrach:2012,Raykar:09}. It is important to note that we are considering a separate, and perhaps more general, problem: we have only one source of noisy labels, and the errors need not come from the human annotators, but could be introduced through contamination or automatic labelling.

\section{Model}
Recall that in binary logistic regression, the probability of an example $x_i$ being positive is modeled as $$g(\theta^Tx_i) = \frac{1}{1 + e^{-\theta^Tx_i}}.$$ For simplicity we assume the intercept term has been folded into the weight vector $\theta$, so $\theta \in \mathbb{R}^{m+1}$ where $m$ is the number of features.

Following ~\newcite{She:11}, we propose the following robust extension: for each datapoint $i=1, \dots, n$, we introduce a real-valued shift parameter $\gamma_i$ so that the sigmoid becomes

$$g(\theta^Tx_i + \gamma_i) = \frac{1}{1 + e^{-\theta^Tx_i - \gamma_i}}$$
Since we believe that most examples are correctly labelled, we $L_1$-regularize the shift parameters to encourage sparsity. Letting $y_i \in \{0, 1\}$ be the label for datapoint $i$ and fixing $\lambda \geq 0$, our objective is now given by
\begin{equation}
l(\theta, \gamma) = \sum_{i=1}^n  y_i \log g(\theta^Tx_i + \gamma_i) + (1 - y_i)\log \left(1 - g(\theta^Tx_i + \gamma_i)\right) -  \lambda \sum_{i=1}^n \abs{\gamma_i}
\end{equation}

These parameters $\gamma_i$ let certain datapoints shift along the sigmoid, perhaps switching from one class to the other. If a datapoint $i$ is correctly annotated, then we would expect its corresponding $\gamma_i$ to be zero. If it actually belongs to the positive class but is labelled negative, then $\gamma_i$ might be positive, and analogously for the other direction.

One way to interpret the model is that it allows the log-odds of select datapoints to be shifted. Compared to models based on label-flipping, where there is a global set of flipping probabilities, our method has the advantage of targeting each example individually.

It is worth noting that there is no difficulty in regularizing the $\theta$ parameters as well. For example, if we choose to use an $L_1$ penalty then our objective becomes
$$l(\theta, \gamma) = \sum_{i=1}^n y_i \log g(\theta^Tx_i + \gamma_i) + (1 - y_i)\log \left(1 - g(\theta^Tx_i + \gamma_i)\right) - \kappa \sum_{j=1}^m\abs{\theta_j} - \lambda \sum_{i=1}^n \abs{\gamma_i} ~~~~ (2) $$

Finally, it may seem concerning that we have introduced a new parameter for each datapoint. But in many applications the number of features already exceeds $n$, so with proper regularization, this increase is actually quite reasonable.

\subsection{Training}
Notice that adding these shift parameters is equivalent to introducing $n$ features, where the $i$th new feature is $1$ for datapoint $i$ and $0$ otherwise. With this observation, we can simply modify the design matrix and parameter vector and train the logistic model as usual. Specifically, we let $\theta' = (\theta_0, \ldots, \theta_m, \gamma_1, \ldots, \gamma_n)$ and $X' = [X | I_n]$ so that the objective simplifies to
\begin{equation}
l(\theta') = \sum_{i=1}^n y_i \log g(\theta'^Tx'_i) + (1 - y_i) \log \left(1 - g(\theta'^Tx'_i)\right) - \lambda \!\!\!\! \sum_{j=m+1}^{m+n} \abs{\theta'^{(j)}} \nonumber
\end{equation}

\noindent Upon writing the objective in this way, we immediately see that it is convex, just as standard $L_1$-penalized logistic regression is convex.

One small complication is that the parameters corresponding to $\gamma$ are now regularized, while those corresponding to $\theta$ are not (or perhaps we wish to regularize them differently). In practice this situation does not pose much difficulty, and in Appendix C we show how to train these models using standard software.

\subsection{Testing}
To obtain our final logistic model, we keep only the $\theta$ parameters. Predictions are then made as usual: $${\bf I}\{g(\hat\theta^Tx) > 0.5\}$$

\subsection{Selecting Regularization Parameters}
The parameter $\lambda$ from equation (1) would normally be chosen through cross-validation. However our set-up is unusual in that the training set may contain errors, and even if we have a designated development set it is unlikely to be error-free. We found in simulations that the errors largely do not interfere in selecting $\lambda$, so in the experiments below we therefore cross-validate as normal. 

Notice that $\lambda$ has a direct affect on the number of nonzero shifts $\gamma$ and hence the suspected number of errors in the training set. So if we have information about the noise level, we can directly incorporate it into the selection procedure. For example, we may believe the training set has no more than 15\% noise, and so would restrict the choice of $\lambda$ during cross-validation to only those values where 15\% or fewer of the estimated shift parameters are nonzero

We now consider situations in which the $\theta$ parameters are regularized as well. Assume, for example, that we use $L_1$-regularization as in equation (2). We would then need to optimize over both $\kappa$ and $\lambda$. In cases like these it is common to first construct a one-dimensional family, so we can then cross-validate a single parameter~\cite{Friedman:09,Arlot:10}. In addition to being faster to compute, this method gives more accurate estimates of the true error rate.

Concretely, we perform the following procedure:
\begin{enumerate}
\item For each $\kappa$ of interest, find the value of $\lambda$ that, along with this choice of $\kappa$, maximizes the robust model's accuracy on the train set.
\item Cross-validate to find the best choice for $\kappa$, using the corresponding values for $\lambda$ found in the first step.
\end{enumerate}

\noindent Note that it is fine to choose $\lambda$ based on training accuracy, since it is not used in making predictions and so there is little risk of overfitting.

For large, high-dimensional datasets even this procedure may be too costly, and training accuracy is not always informative. So in the natural language processing experiments below, we adopt a simpler strategy:
\begin{enumerate}
\item Cross-validate using standard logistic regression to select $\kappa$.
\item Fix this value for $\kappa$, and cross-validate using the robust model to find the best choice of $\lambda$.
\end{enumerate}
Although not as well-motivated theoretically, this method still produces good results.

\section{Experiments}

We now present several experiments to assess the effectiveness of the approach, ranging from simulations in which labels are flipped uniformly at random, to experiments on natural language datasets where annotation errors are quite systematic. These experiments measure the robust model against standard logistic regression; for a comparison with other methods for handling annotation errors, please see Appendix B.

\subsection{Simulated Data}

In our first experiment, we simulate logistic data with 10 features drawn Uniform(-5, 5), letting $\theta_j = 2$ for $j=1, \ldots, m$ and the intercept be zero. We create training, development, and test sets containing 500 examples each and introduce noise into both the training and development sets by flipping labels uniformly at random. The regularization parameter $\lambda$ is chosen simply by minimizing 0-1 loss on the development set. For all simulation experiments we use \texttt{glmnet}, an R package that trains both lasso~($L_1$)-penalized and elastic net models through cyclical coordinate descent~\cite{Friedman:09}. The results for standard versus robust logistic regression are shown in Table 1, for various levels of noise.

Using the tuning procedure described in Section 3.3, we next perform simulations in which the original features are $L_1$-penalized as well (see Table 1). We generate logistic data with 20 features, only 5 of which are relevant, and again set  $\theta_j = 2$ for $j=1, \ldots, m$ and the intercept to zero. The training, development, and test sets are each of size 100, and label noise is added to all data but the test set. The regularization parameter for the baseline model is tuned on the development set. Additional implementation details can be found Appendix C.

\begin{table}
\tabcolsep=6pt
\centering
\begin{tabular} {c c c c c}
$p_0$ & $p_1$ & regularized & standard & robust  \\
\hline
0.0 & 0.0 & no & $96.56 \pm 0.09$ & $96.60 \pm 0.10$ \\
0.1 & 0.0 & no & $93.48 \pm 0.18$ & $93.58 \pm 0.18$ \\
0.2 & 0.0 & no & $87.49 \pm 0.24$ & ${\bf 89.22 \pm 0.23}$ \\
0.3 & 0.0 & no & $80.40 \pm 0.25$ & ${\bf 84.15 \pm 0.28}$ \\
0.3 & 0.1 & no &  $84.16 \pm 0.35$ & ${\bf 86.63 \pm 0.33}$ \\  
0.3 & 0.0 & yes & $75.89 \pm 0.50$ & ${\bf 76.98 \pm 0.56}$ \\
0.3 & 0.1 & yes & $74.98 \pm 0.56 $& ${\bf 76.16 \pm 0.57}$ \\  
\end{tabular}
\caption{Accuracy of standard vs. robust logistic regression for various levels of noise. The $p_0$ column gives the probability of class 0 flipping to 1, and vice versa for $p_1$.}
\vspace{-2.0cm}
\end{table}

\begin{figure}
\vspace{-1.0cm}
\centering
\includegraphics[width=10cm]{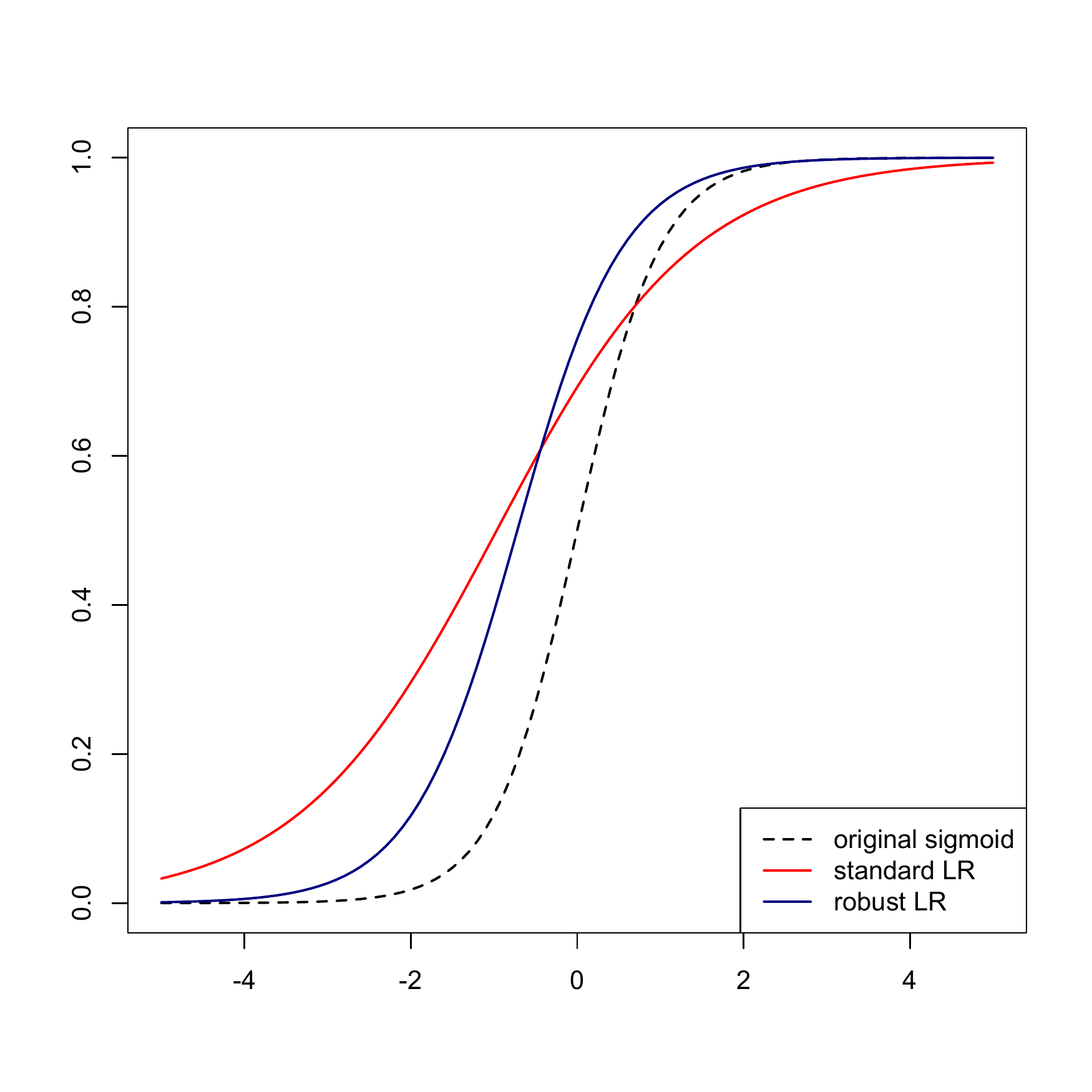}
\vspace{-1.0cm}
\caption{Fit resulting from a standard vs. robust model, where data is generated from the dashed sigmoid and negative labels flipped with probability 0.2.}
\end{figure}

As the results show, robust logistic regression provides a consistent improvement over the baseline. The performance difference grows larger with the amount of label noise, and is also evident when labels have been flipped in both directions.  A one-dimensional example of this improvement is seen in Figure 2.

Importantly, the model does not perform worse than standard logistic regression when no errors are present. Inspecting the learned parameters from these runs, we see that almost all $\gamma$ have been set to 0.

\subsection{Contaminated Data}
We next apply our approach to a biological dataset with suspected labelling errors. Called the colon cancer dataset, it contains the expression levels of 2000 genes from 40 tumor and 22 normal tissues~\cite{Alon:99}. There is evidence in the literature that certain tissue samples may have been cross-contaminated. In particular, 5 tumor and 4 normal samples should have their labels flipped.

Since the dataset is so small, it is difficult to accurately measure the performance of our model against the baseline. We instead examine its ability to identify mislabelled training examples. Because there are many more features than datapoints and it is likely that not all genes are relevant, we choose to place an $L_1$ penalty on $\theta$. 

Using \texttt{glmnet}, we again select $\kappa$ and $\lambda$ using the cross-validation procedure from Section 3.3. Looking at the resulting values for $\gamma$, we find that only 7 of the shift parameters are nonzero and that each one corresponds to a suspicious datapoint. As further confirmation, the sign of the gammas correctly match the direction of the mislabelling. Compared to previous attempts to automatically detect errors in this dataset, our approach identifies at least as many suspicious examples but with no false positives. A detailed comparison is given in Table 2. Although ~\newcite{Bootkrajang:12} are quite accurate, it is worth noting that due to its nonconvexity, their model needed to be trained 20 times to achieve these results.

\begin{table}
\tabcolsep=3pt
\noindent\makebox[\textwidth]{
\begin{tabularx}{1.1\textwidth}{l | c c c c c c c c c | l }
 \multicolumn{1}{c}{method} & \multicolumn{9}{| c |}{suspects identified} & \multicolumn{1}{c}{false positives} \\
\hline
\newcite{Alon:99} & T2 & T30 & T33 & T36 & T37 & N8 & N12 & N34 & N36 & \\
\newcite{Furey:00} & & $\bullet$ & $\bullet$ & $\bullet$ & & $\bullet$ & & $\bullet$ & $\bullet$ & \\
\newcite{Kadota:03} & $\bullet$ & & & & $\bullet$ & $\bullet$ & & $\bullet$ & $\bullet$ &T6, N2 \\
\newcite{Malossini:06} & $\bullet$ & $\bullet$ & $\bullet$ & $\bullet$ & & & $\bullet$ & $\bullet$ & $\bullet$ & T8, N2, N28, N29 \\
Bootkrajang et al. (2012) & $\bullet$ & $\bullet$ & $\bullet$ & $\bullet$ & & & $\bullet$ & $\bullet$ & $\bullet$ & \\
\hline
robust LR & & $\bullet$ & $\bullet$ & $\bullet$ & & $\bullet$ & $\bullet$ & $\bullet$ & $\bullet$ & 
\end{tabularx}}
\caption{Results of various error-identification methods on the colon cancer dataset. The first row lists the samples that are biologically confirmed to be suspicious, and each other row gives the output from an automatic detection method. Bootkrajang et al. report confidences, so we threshold at 0.5 to obtain these results.}
\end{table}

\subsection{Manually Annotated Data}

In these experiments we focus on a classic task in NLP called \emph{named entity recognition}. In the traditional set-up, the goal is to determine whether each word is a person, organization, location, or not a named entity (`other'). Since our model is binary, we concentrate on the task of deciding whether a word is a person or not. This task does not trivially reduce to finding the capitalized words, as the model must distinguish between people and other named entities like organizations.

For training, we use a large, noisy NER dataset collected by Jenny Finkel. The data was created by taking various Wikipedia articles and giving them to five Amazon Mechanical Turkers to annotate. Few to no quality controls were put in place, so that certain annotators produced very noisy labels. To construct the train set we chose a Turker who was about average in how much he disagreed with the majority vote, and used only his annotations. Negative examples are subsampled to bring the class ratio to a reasonable level (around 1 to 10). We evaluate on the development test set from the CoNLL shared task \cite{Sang:03}. This data consists of news articles from the Reuters corpus, hand-annotated by researchers at the University of Antwerp. More details about the dataset can be found in Table 3.

\begin{table*}
\centering
\tabcolsep=5pt
\begin{tabular} {c c c c c c c}
source & partition & \# pos & \# neg & \# features & $p_0$ & $p_1$ \\
\hline
CoNLL + MUSE & train & 8,392 & 80,000 & 190,185 & 0.371 & 0.007 \\
Wikipedia & train & 24,002 & 200,000 & 393,633 & 0.004 & 0.075 \\
CoNLL & test & 3,149 & 48,429 & 125,062 & - & -\\
\end{tabular}
\caption{Statistics about the data used in the NER experiments. For the Wikipedia train set, the $p_0$ column represents the fraction of examples that the majority agreed were negative, but that the chosen annotator marked positive (and analogously for $p_1$). We still include examples for which there was no majority consensus, so these noise estimates are quite conservative. As for the MUSE data, the $p_0$ column gives the fraction of examples that are marked positive in the official CoNLL train set, but that the automatic system labelled negative, and vice versa for $p_1$.}
\end{table*}

We extract a set of features using Stanford's NER pipeline \cite{Finkel:05}. This set was chosen for simplicity and is not highly engineered -- it largely consists of lexical features such as the current word, the previous and next words in the sentence, as well as character n-grams and various word shape features. We choose to $L_2$-regularize the features, so that our penalty now becomes $$\frac{1}{2\sigma^2} \sum_{j=0}^m \abs{\theta_j}^2 + \lambda \sum_{i=1}^n \abs{\gamma_i}$$ This choice is natural as $L_2$ is the most common form of regularization in NLP, and we wish to verify that our approach works for penalties besides $L_1$.

\begin{figure}
\centering
\includegraphics[width=10cm]{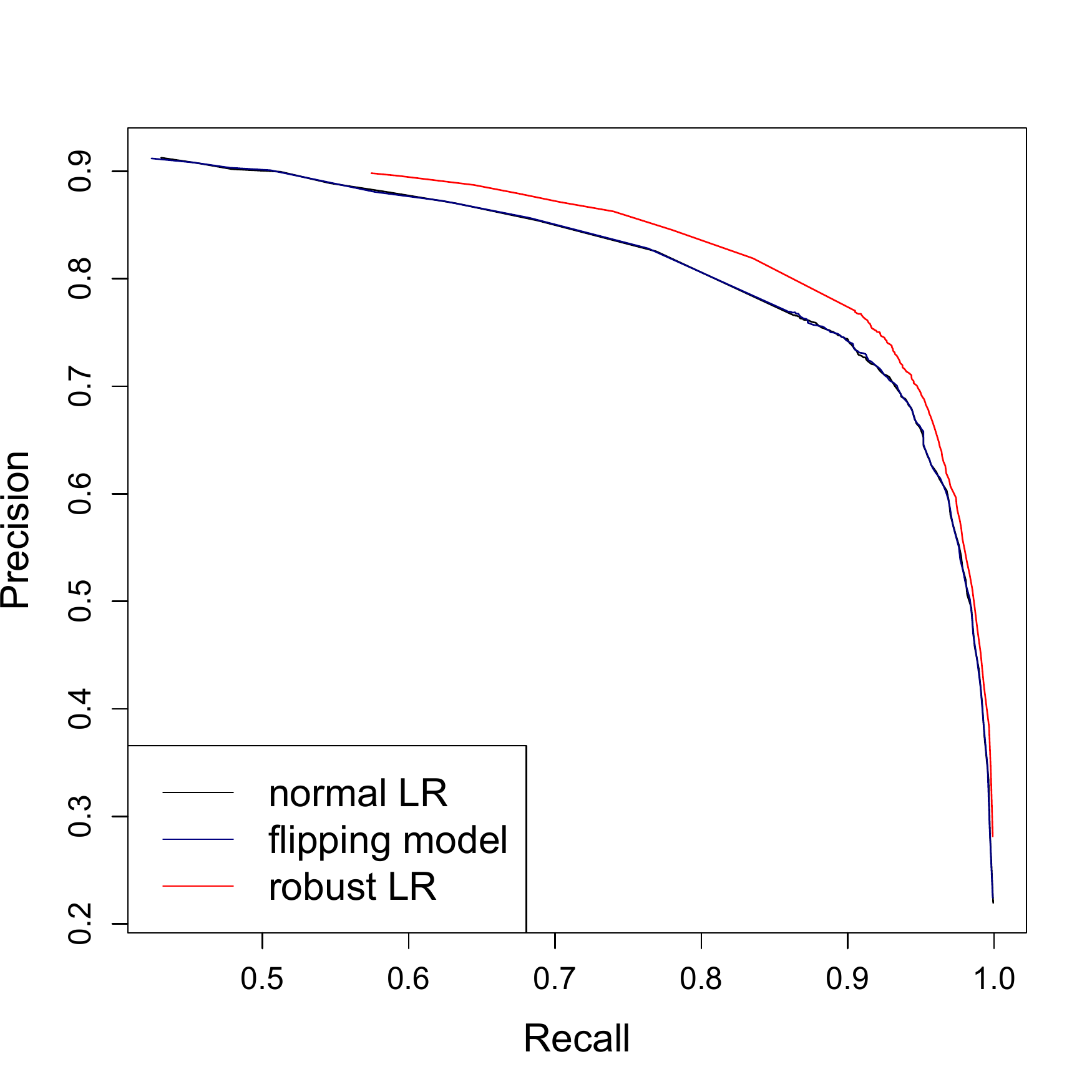}
\caption{Precision-recall curve obtained from training on noisy Wikipedia data and testing on CoNLL. The flipping model refers to the approach from Bootkrajang and Kaban (2012).}
\end{figure}

\begin{table}
\centering
\begin{tabular} {c c c c}
model & precision & recall & F1 \\
\hline
standard & 76.99 & 85.87 & 81.19 \\
flipping & 76.62 & 86.28 & 81.17 \\
robust & {\bf 77.04} & {\bf 90.47} & {\bf 83.22} \\
\end{tabular}
\caption{Performance of standard vs. robust logistic regression in the Wikipedia NER experiment. The flipping model refers to the approach from Bootkrajang and Kaban (2012).}
\vspace{-0.2cm}
\end{table}

The robust model is fit using Orthant-Wise Limited-Memory Quasi Newton (OWL-QN), a technique for optimizing an $L_1$-penalized objective \cite{Andrew:07}. We tune both models through 5-fold cross-validation to obtain $\sigma^2 = 1.0$ and $\lambda = 0.1$. Note that from the way we cross-validate (first tuning $\sigma$ using standard logistic regression, fixing this choice, then tuning $\lambda$) our procedure may give an unfair advantage to the baseline.

We also compare against the algorithm proposed in ~\newcite{Bootkrajang:12}, an extension of logistic regression mentioned in the section on prior work. This model assumes that annotation errors are produced from label-flipping: each example's true label is flipped with a certain probability before being observed. The features are linked to the latent `true' labels through a standard logistic classifier, and these labels relate to the observed ones through two global parameters, the probability of a positive label flipping to negative, and the probability of a negative label flipping to positive. The model is trained by first estimating the latent `true' labels, then learning the weights of the logistic classifier and the values for the two flipping probabilities. During testing, these flipping probabilities are discarded and predictions are made using only the logistic classifier.

The results of these experiments are shown in Table 4 as well as Figure 3. Robust logistic regression offers a noticeable improvement over the baseline, and this improvement holds at essentially all levels of precision and recall. Interestingly, the flipping model show no substantial difference with standard logistic regression. A more in-depth discussion of this outcome is given in Section 5.

\subsection{Automatically Annotated Data}

We now turn to a setting in which training data has been automatically generated. The task is the same as in the previous experiment: for each word in a sentence we must identify whether it represents a person or not. For evaluation we again use the development test set from the CoNLL shared task, and extract the same set of simple features as before.

As for the training data, we take the sentences from the official CoNLL train set and run them through a simple NER system to create noisy labels. We use a system called MUSE, which makes use of gazetteers and hand-crafted rules to recognize named entities \cite{Maynard:01}. The software is distributed with GATE, a general purpose set of tools for processing text, and is not tuned for any particular corpus \cite{Cunningham:02}. We have again subsampled negatives to achieve a ratio of roughly 1 to 10. More information about the data can be found in Table 5. Somewhat expectedly, we see that the system has a high false negative rate.

We again use 5-fold cross-validation to tune the regularization parameters, ultimately picking $\sigma^2 = 10$ and $\lambda = 0.01$. Our first attempt at selecting $\lambda$ gave a very large value, so that nearly all of the resulting $\gamma$ parameters were zero. We therefore decided to use our knowledge of the noise level to guide the choice of regularization. In particular, we restrict our choice of $\lambda$ so that the proportion of $\gamma$ parameters which are nonzero roughly matches the fraction of training examples that are mislabelled (around 4\%, after summing across both classes). Note that even in more realistic situations, where expert labels are not available, we can often gain a reasonable estimate of this number.

\begin{table}
\centering
\begin{tabular} {c c c c}
model & precision & recall & F1 \\
\hline
standard & 84.52 & 70.91 & 77.12 \\
flipping & 84.36 & 70.91 & 77.05 \\
robust & {\bf 84.64} & {\bf 72.44} & {\bf 78.06} \\
\end{tabular}
\caption{Performance of standard vs. robust logistic regression in the Muse NER experiment. The flipping model refers to the approach from Bootkrajang and Kaban (2012).}
\end{table}

Table 5 shows the experimental results. We see that on this dataset robust logistic regression offers a modest improvement over the baseline. The flipping model again behaves nearly identically to standard logistic regression.

\section{Discussion}
In the simulation experiments from Section 4.1, the robust model offers a notable advantage over the baseline if the features are uniformly distributed. But when we rerun the experiments with features drawn Normal(0, 1), the improvement in accuracy decreases by as much as 1\%. One explanation is as follows: in this situation, the datapoints, and therefore the annotation errors, tend to cluster around the border between positive and negative. Logistic regression, by virtue of its probabilistic assumptions, is naturally forgiving toward points near its decision boundary. So when noise is concentrated at the border, adding shift parameters does not provide the same benefit. In short, the robust model seems to perform best when there is a good number of mislabelled examples that are not close cases.

As noted in the NER experiments, the robust model shows less of an improvement when the training data is generated automatically rather than manually. One likely explanation is that more than human annotators, rule-based systems tend to make mistakes on examples with similar features. For example if a certain word was not in MUSE's gazetteers, and so it incorrectly labelled every instance of this word as negative, we might have a good number of erroneous examples that are close together in feature space. In this setting it can be hard for any robust classifier to learn what is mislabelled.

We also observe that the flipping model performs essentially the same as standard logistic regression. During training, the two flipping probabilities consistently converge to 0, which corresponds to the situation in which no label-flipping occurred. Learning the weights for the logistic classifier then gives exactly the same values as would standard logistic regression. Updates to the way the parameters were initialized, including several attempts at randomization, failed to change this outcome. 

A likely explanation is that given the large ratio of features to datapoints so common in NLP applications, the classifier's weights already provide more than enough degrees of freedom, and so the model essentially ignores the extra flipping parameters. When an example is mislabelled, it is likely better to `fiddle' one of the many weights instead of modifying a global probability, which has major repercussions across examples. Neither strengthening the $L_2$-regularization nor even switching to an $L_1$ penalty helped the probabilities converge to a nonzero value. Our model manages to avoid this issue by introducing one shift parameter per datapoint. The $\gamma$ variables allow for fine-grained corrections, and have a large enough presence to compete with the classifier's weights.

\section{Comparison to SVMs}
It is interesting to observe the similarity between this model and a soft-margin SVM:
\begin{align}
&\min_{w, \xi, b} ~ \frac{1}{2} \norm w^2 + C \sum_{i=1}^n \xi_i \nonumber \\
&\text{s. t.} ~~ \forall_i ~~ y_i(w^Tx_i - b) \geq 1 - \xi_i, ~~~ \xi_i \geq 0 \nonumber
\end{align}
The $\gamma$ parameters correspond to slack variables $\xi_i$, which allow certain datapoints to lie on the wrong side of the separating hyperplane. As in our model, these slack variables are $L_1$-penalized to promote sparsity. One very reasonable interpretation of our approach is that we've added slack variables to logistic regression, and much as they help robustify SVMs, slack variables can benefit GLMs as well.

However, it is important to remember that these approaches have significant differences, and can have widely varying performance in practice. Take for example the following simulation, where positive and negative examples are drawn from two distinct multivariate normals, with ${\bf \mu} = (0, 0), {\bf \Sigma} = 1 \cdot I_2$ and ${\bf \mu} = (1, 1), {\bf \Sigma} = 1.5 \cdot I_2$, respectively. On a clean dataset, the robust model achieves an accuracy of $73.33 \pm 0.41\%$, and a soft-margin SVM performs very similarly, at $73.28 \pm 0.42\%$. After flipping negative labels uniformly at random with probability 0.3, the robust model has performs at $66.37 \pm 0.60\%$, while the SVM drops to $63.81\pm 0.85\%$.

\section{Future Work}
A natural direction for future work is to extend the model to a multi-class setting. One option is to introduce a $\gamma$ for every class except the negative one, so that there are $n(c-1)$ shift parameters in all. We could then apply a group lasso, with each group consisting of the $\gamma$ for a particular datapoint \cite{Meier:08}. This way all of a datapoint's shift parameters drop out together, which corresponds to the example being correctly labelled. A simpler approach is to use one-vs-all classification: we train one binary robust model for each class, and have them vote on an example's label. We have found preliminary success with this method in a relation extraction task.  

CRFs and other sequence models could also benefit from the addition of shift parameters. Since the extra variables can be neatly folded into the linear term, convexity is preserved and the model could essentially be trained as usual.

\section{Conclusion}
We presented a robust extension of binary logistic regression that can outperform the standard model when annotation errors are present. Our method introduces shift parameters to allow datapoints to move across the decision boundary. It largely maintains the efficiency and scalability of logistic regression, but is better equipped to train with noisy data and can also help identify mislabelled examples. 

As large, noisy datasets continue to gain prevalence, it is important to develop classifiers with robustness in mind. Most promising seem to be models that incorporate the potential for mislabelling as they train. We presented one such model, and demonstrated that explicitly accounting for annotation errors can provide significant benefit. 

\section*{Acknowledgments} I am very grateful to my advisor Chris Manning for being so encouraging over the past few years, and for his many helpful insights and suggestions. Thank you to the whole Stanford NLP group for being so welcoming during my time as an undergraduate and Masters student. I would especially like to thank Mihai Surdeanu for being a patient and encouraging mentor, and Gabor Angeli for his suggestions.

Finally, I am thankful to Rob Tibshirani and Stefan Wager for their invaluable advice and support.

\newpage
\section*{Appendix A. The Label-Flipping Model}
Here we give a more careful description of the model from ~\cite{Bootkrajang:12}, and fill in details missing from the original presentation. This model assumes that annotations errors are produced from label-flipping: each example's true label is flipped with a certain probability before being observed. Our notation in the derivations is largely the same as that introduced in Section 3.

\subsection*{Model}
The authors modify standard logistic regression to contain latent variables $z$ representing a `true label' for each datapoint. Then $x$ relates to $z$ through a logistic model as usual, and $z$ is connected to the observed label $y$ through a collection of flipping probabilities $\gamma$. There is one $\gamma$ per pair of classes, so that $\gamma_{ab}$ represents the probability of an example's label flipping from class $a$ to $b$. Figure 3 represents this set-up as a graphical model. During testing, we discard the $\gamma$ parameters and predict using only $\theta$. 

\subsection*{EM derivation}
The authors present an iterative algorithm for learning $\theta$ and $\gamma$, but we find it simpler and more informative to use Expectation Maximization, a common method for estimating parameters in latent-variable models. To begin, the log-likelihood of the data is given by
$$ l(\theta, \gamma) = \sum_{i=1}^n \log p(y^{(i)} | x^{(i)}, \theta, \gamma) = \sum_{i=1}^n \log \sum_z p(y^{(i)}, z | x^{(i)}, \theta, \gamma)$$
In EM we work with the expected log-likelihood of the joint, which provides a convex lower bound on the true likelihood:
$$ Q(\theta, \gamma) = \sum_{i=1}^n \sum_z p(z | y^{(i)}, x^{(i)}) \log p(y^{(i)}, z | x^{(i)}, \theta, \gamma) $$
\paragraph{E-step} Infer the latent distribution $p(z|y, x)$. Letting $g$ denote the sigmoid function, $\theta^{(c)}$ represent the weight vector for class $c$ and $\gamma_{c_1c_2}$ be the probability of flipping from class $c_1$ to $c_2$, we have
$$ p(c|y, x) = \frac{p(y, c|x)}{p(y|x)} = \frac{p(y|c)p(c|x)}{\sum_z p(y|z)p(z|x)}  = \frac{\gamma_{cy} g(x^{T}\theta^{(c)})}{\sum_z \gamma_{zy}g(x^{T}\theta^{(z)})}$$

\begin{figure}
\centering
\vspace{-2.5cm}
\includegraphics[width=12cm]{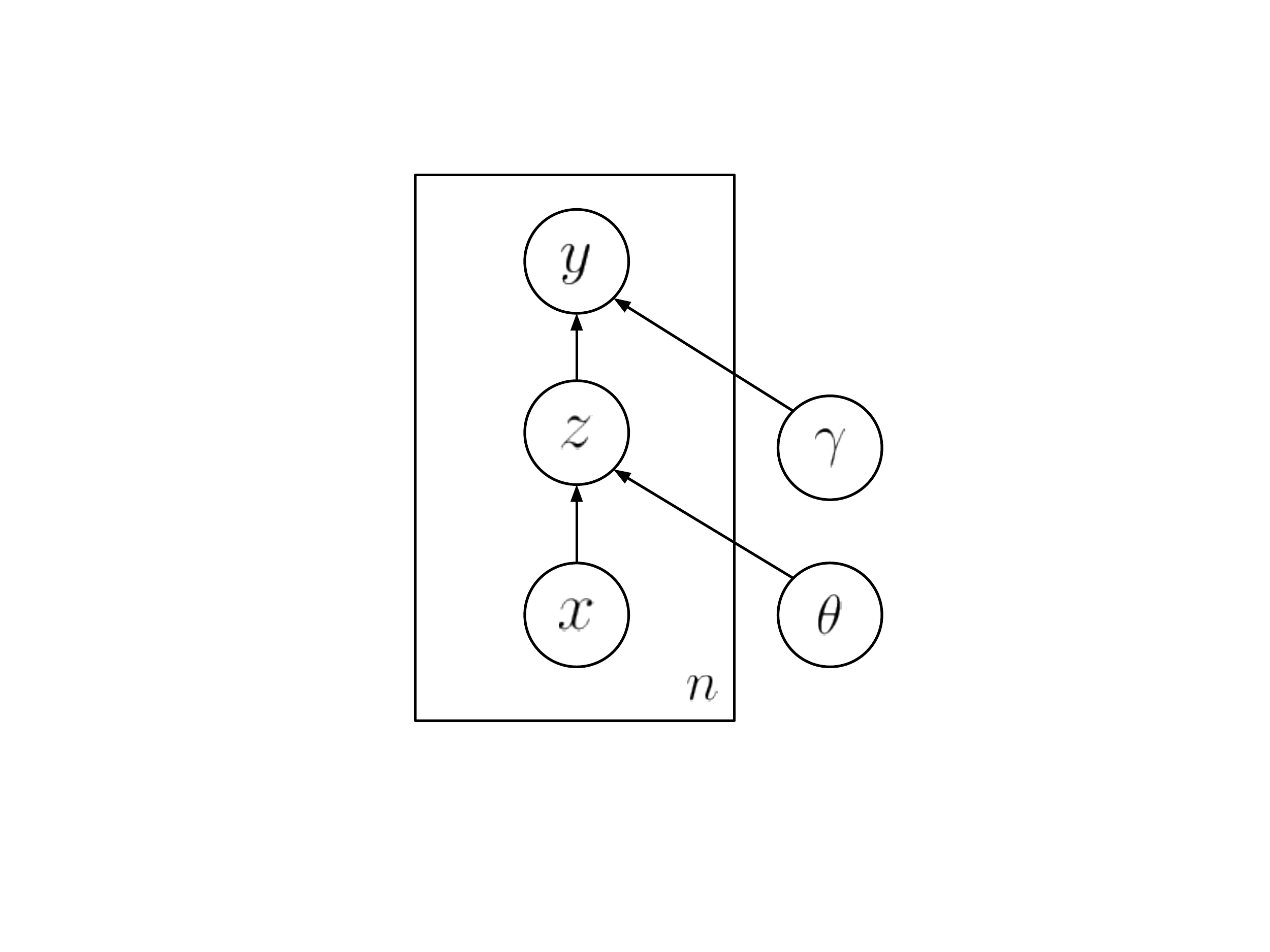}
\vspace{-1.2cm}
\caption{Plate diagram for the label-flipping model.}
\end{figure}

\paragraph{M-step} Maximize the expected log-likelihood to obtain $\gamma$ and $\theta$. Using the definition of $Q$ above we have 
\begin{eqnarray}
  Q(\theta, \gamma) &=& \sum_{i=1}^n \sum_z p(z | y^{(i)}, x^{(i)}) \log \left( p(y^{(i)}|z, \gamma)p(z | x^{(i)}, \theta) \right) \nonumber \\
   &=& \sum_{i=1}^n \sum_z p(z | y^{(i)}, x^{(i)}) \log \left( \gamma_{zy}~g(x^{(i)T}\theta^{(z)}) \right) \nonumber \\
   &=& \sum_{i=1}^n \sum_z p(z | y^{(i)}, x^{(i)}) \log \gamma_{zy} + \sum_{i=1}^n \sum_z p(z | y^{(i)}, x^{(i)}) \log g(x^{(i)T}\theta^{(z)})~~~~~~~
\end{eqnarray}
It is clear from this equation that $\gamma$ and $\theta$ can be maximized separately. For the $\gamma$ parameters, we have the additional constraint that for fixed $c$, the $\gamma_{cz}$ must sum to $1$. We construct the Lagrangian and set derivatives to zero:
\begin{eqnarray}
  Q(\gamma) &=& \sum_{i=1}^n \sum_z p(z | y^{(i)}, x^{(i)}) \log \gamma_{zy^{(i)}} + \sum_{z_1}\beta_{z_1}(1 - \sum_{z_2}\gamma_{z_1z_2}) \nonumber \\
    0 &=& \frac{\partial Q(\gamma)}{\partial \gamma_{c_1c_2}} = \sum_{i=1}^n I\{y^{(i)} = c_2\}~\frac{p(c_1|c_2, x^{(i)})}{\gamma_{c_1c_2}} + \beta_{c_1}  \nonumber
\end{eqnarray}
which implies
\begin{eqnarray}
\gamma_{c_1c_2} = \frac{\sum_i I\{y^{(i)} = c_2\}~p(c_1| y^{(i)}, x^{(i)}) }{-\beta_{z_1}} = \frac{\sum_i^n I\{y^{(i)} = c_2\}~p(c_1|y^{(i)}, x^{(i)})}{\sum_i p(c_1| y^{(i)}, x^{(i)})}~~~~
\end{eqnarray}
Now for the $\theta^{(c)}$ we first calculate
\begin{eqnarray}
\frac{\partial}{\partial \theta_j^{(c_1)}} g(x^{T}\theta^{(c_1)}) &=&  \frac{\partial}{\partial \theta_j^{(c_1)}} \left( \frac{e^{-x^{T}\theta^{(c_1)}}}{1 + \sum_c e^{-x^{T}\theta^{(c)}}} \right) \nonumber \\ 
&=&  \frac{-x_j e^{-x^{T}\theta^{(c_1)}}}{1 + \sum_c e^{-x^{T}\theta^{(c)}}} -  \frac{e^{-x^{T}\theta^{(c_1)}}}{(1 + \sum_c e^{-x^{T}\theta^{(c)}})^2} \cdot -x_j e^{-x^{T}\theta^{(c_1)}} \nonumber \\
&=& \left( g(x^{T}\theta^{(c_1)})^2 - g(x^{T}\theta^{(c_1)}) \right) x_j \nonumber
\end{eqnarray}
Through a similar process, we can derive
\begin{eqnarray}
\frac{\partial}{\partial \theta_j^{(c_1)}} g(x^{T}\theta^{(c_2)}) &=& g(x^{T}\theta^{(c_1)})g(x^{T}\theta^{(c_2)}) x_j \nonumber
\end{eqnarray}

\noindent and for $c_2 = 0$,
\begin{eqnarray}
\frac{\partial}{\partial \theta_j^{(c_1)}}  \frac{1}{1 + \sum_c e^{-x^{T}\theta^{(c)}}} &=& g(x^{T}\theta^{(c_1)}) \left( \frac{1}{(1 + \sum_c e^{-x^{T}\theta^{(c)}})^2} \right) x_j \nonumber
\end{eqnarray}

We are now in a position to calculate the gradient with respect to $\theta_i^{(c_1)}$:
\begin{eqnarray}
\frac{\partial Q(\theta)}{\partial \theta_j^{(c_1)}} &=& \sum_{i=1}^n \sum_z p(z|y^{(i)}, x^{(i)}) \frac{\partial}{\partial \theta_j^{(c_1)}} \log g(x^{(n)T}\theta^{(z)}) \nonumber \\
&=& \sum_{i=1}^n \left(\sum_{z \neq c_1} p(z|y^{(i)}, x^{(i)})g(x^{(i)T}\theta^{(c_1)})  + p(c_1|y^{(i)}, x^{(i)})(g(x^{(i)T}\theta^{(c_1)}) - 1 ) \right)  x_j^{(i)} \nonumber \\
&=& \sum_{i=1}^n \left( g(x^{(i)T}\theta^{(c_1)}) \sum_z p(z|y^{(i)}, x^{(i)}) - p(c_1|y^{(i)}, x^{(i)}) \right)  x_j^{(i)} \nonumber \\
&=& \sum_{i=1}^n \left( g(x^{(i)T}\theta^{(c_1)}) - p(c_1|y^{(i)}, x^{(i)}) \right) x_j^{(i)}
\end{eqnarray}

Equation (3) is quite intuitive -- we obtain nearly the same gradient as in standard multinomial logistic regression 
$$\frac{\partial Q(\theta)}{\partial \theta_j^{(c_1)}}  =  \sum_{i=1}^n \left( g(x^{(i)T}\theta^{(c_1)}) - I\{z^{(i)} = c_1\} \right) x_j^{(i)} $$ 
except that $I\{z^{(i)} = c_1\}$ has been replaced with its expectation, $p(c_1|y^{(i)}, x^{(i)})$.

\subsection*{Connection to Instance-Weighting}
Interestingly, we can cast the above procedure for selecting $\theta$ as a form of instance-weighting. Concretely, we copy every datapoint $k$ times where $k$ is the number of classes, so that each copy corresponds to a possible class. Then by weighting copy $c$ by $p(c|y^{(i)}, x^{(i)})$, we recover the second term from equation (1), which is what we set out to optimize.

Standard instance-weighting techniques determine the class-membership probabilities $p(c|y^{(i)}, x^{(i)})$ using an example's distance from the centroid of each class~\cite{Rebbapragada:07,Thiel:08}. The label-flipping model, in contrast, iteratively estimates these probabilities through EM. While this model is in some ways more sophisticated, previous instance-weighting methods have shown to be effective empirically and may still provide important insight. In particular, instead of assuming that each datapoint flips from class $a$ to $b$ with fixed probability $\gamma_{ab}$, we could define $\gamma_{ab}$ be a function (perhaps sigmoid) of the datapoint's distance to the class centroids. This way the flipping probabilities can be more fine-grained, encoding information specific to each datapoint.

\section*{Appendix B. Comparison with Other Methods}
We now compare the robust model against other methods designed to handle annotation errors. In particular, we test its performance against the model from ~\newcite{Bootkrajang:12}, which is a good representative of approaches based on a label-flipping model. We also compare against a common form of prefiltering based on k-nearest neighbors that was popularized by~\newcite{Brodley:99}.

\subsection*{Simulation Experiments}
Unless noted otherwise, in all experiments that follow we simulate logistic data with 50 features drawn Uniform(-5, 5), with $\theta_j = 2$ for $j = 1, \ldots, m$ and zero intercept. We create training, development, and test sets each of size 500, and noise is introduced into the training and development sets by flipping negative labels uniformly at random with $0.3$ probability.

The filtering approach is implemented as follows: for each example, we examine its $k$ nearest neighbors in feature space. If the label of one of these neighbors disagrees with the example's label, then it is discarded from the train set. A logistic classifier is then trained on this filtered dataset. We select $k$ based on the development set. Note that if $k$ is 1, then no examples are filtered and the model falls back to standard logistic regression. We implement the label-flipping model through the EM procedure derived in Appendix A, learning the weights for $\theta$ using a standard package for logistic regression that supports instance weighting. The results of these simulations can be found in Table 7.

\paragraph{Experiment 1} In our first experiment, errors are introduced by flipping negative labels uniformly at random with $0.3$ probability. While robust logistic regression does provide an improvement over the baseline, the prefiltering and flipping models perform substantially better. The assumptions of this simulation exactly match those of the flipping model, so it naturally performs well. (As further evidence the flipping model achieves a good fit, it correctly learns the entries of the $\gamma$ matrix.) Prefiltering also achieves an impressive accuracy, as the simulation matches the model's assumption that an example's nearest neighbors should share its label.

\paragraph{Experiment 2} We next simulate data with features drawn Normal(0, 1). The results are similar to those in the first experiment. We see that prefiltering is still able to correctly identify mislabelled examples, although it now selects smaller values for $k$ since many datapoints lie near the decision boundary. The robust and flipping models show somewhat less of an improvement over the baseline, perhaps because it is harder to distinguish the points that are truly mislabelled from those that have crossed the boundary by chance.

\begin{table}
\centering
\begin{tabular} {c c c c c}
experiment & baseline & prefiltering & flipping & robust \\
\hline
1 & $79.55 \pm 0.27$ & $87.00 \pm 0.37$ & ${\bf 87.14} \pm 0.13$ & $81.11 \pm 0.27$ \\
2 & $79.15 \pm 0.27$ & ${\bf 86.59} \pm 0.35$ & $85.00 \pm 0.12$ & $80.58 \pm 0.27$ \\
3 & $87.11 \pm 0.25$ & $87.11 \pm 0.25$ &  $91.15 \pm 0.16$ & ${\bf 91.65} \pm 0.14$ \\
4 & $85.95 \pm 0.33$ & $92.87 \pm 0.56$ & $90.28 \pm 0.27$ & ${\bf 94.10} \pm 0.52$ 
\end{tabular}
\caption{Accuracy of each model in the various simulation experiments. The bolded entries highlight the best-performing model.}
\end{table}

\paragraph{Experiment 3} We now set $m=1$ and return to drawing features Uniform(-5, 5) with $\theta= 2$ and zero intercept. In this experiment, errors are introduced in a more systematic way: all negative examples with feature values between -5 and -4 are switched to positive. This set-up represents a plausible scenario in which datapoints with similar feature values are likely to be flipped together. As discussed in Section 5.2, such a situation could arise if the annotations were generated through a noisy automatic process.

We observe during training that prefiltering always reverts to $k=1$, so that it does not filter out any points and performs identically to the baseline. This simulation demonstrates the drawbacks of training the filtering classifier on noisy data. The mislabelled examples are clustered together in feature space, so for most of these points the kNN classifier fails to recognize their labels have been flipped. Moreover, for some values of $k$ the classifier begins to filter out correct points adjacent to the flipped region.

The label-flipping model still performs well, but shows much less of an improvement than in experiments 1 and 2. We suspect that the farther the distribution of errors is from uniform, the less the model's advantage will be.

\paragraph{Experiment 4} In this experiment we test whether these methods are sensitive to model misspecification by generating data that is not quite logistic. Recall that if $p(x|y=0)$ and $p(x|y=1)$ are multi-variate Gaussians with the same covariance matrix, then $p(y|x)$ follows the logistic model. If the two Gaussians have different covariance matrices, then $p(y|x)$ will differ slightly from logistic. Concretely, we still simulate data with features drawn Uniform(-5, 5), but now generate labels using a $\mathcal{N}(-2, 2I)$ distribution for the negative class and $\mathcal{N}(2, I)$ for positive.

All three models perform better than baseline, which demonstrates they are robust to small deviations from the logistic assumption. Interestingly, the robust model shows the most improvement, perhaps suggesting it is the best choice for realistic data.

\subsection*{Natural Language Experiments}
We now run each model on the AMT NER dataset from Section 4.3. For all models the $\theta$ parameters are $L_2$-regularized, and hyper-parameters are selected by cross-validating on the training data. The results are shown in Table 8.
\begin{table}
\centering
\begin{tabular} {c c c c}
model & precision & recall & F1 \\
\hline
standard & 76.99 & 85.87 & 81.19 \\
prefiltering &  76.46 & 90.85 & 83.04 \\
flipping & 76.71 & 86.28 & 81.21 \\  
robust & 77.04 & 90.47 & 83.22
\end{tabular}
\caption{Accuracy of each model in the various simulation experiments. The bolded entries highlight the best-performing model.}
\end{table}

The robust model provides a significant improvement, while the flipping model performs almost identically to the baseline. This result has a simple explanation: during training we observed that the $\gamma$ matrix consistently converged to the identity, so that for each class $c$, $\gamma_{cc} = 1$, but all other components were zero. This solution corresponds to the situation in which no label-flipping occurred, and so gives the same values for $\theta$ as standard logistic regression. Changing the way in which $\gamma$ and $\theta$ are initialized did not seem to help.

We tried randomly subsampling the features to achieve a smaller ratio of features to training examples. With fewer features, the model indeed learns a $\gamma$ matrix different from the identity and gives slightly better accuracy than a baseline, although both models perform very poorly. This finding suggests that in our original experiment, the $\theta$ parameters already provide enough degrees of freedom, so the model essentially ignores the extra $\gamma$ parameters. When an example is mislabelled, it may be better to `fiddle' one of the many $\theta$ parameters, instead of modifying a $\gamma$, which has major repercussions across examples. Neither strengthening the $L_2$-regularization nor even switching to an $L_1$ penalty appeared to help.

Interestingly, the prefiltering approach performs nearly as well as the robust method on this dataset. Looking at the examples it chooses to discard, most of them are indeed misannotated. We suspect that prefiltering succeeds here because the training data is highly redundant. For example the model threw out the word `Wilson', which was incorrectly marked as negative, and we saw many other places in the corpus where `Wilson' had a positive label. In datasets where there is less redundancy, and especially those situations in which errors are systematic, we expect the prefiltering approach to perform worse.

While prefiltering ultimately helps performance, it still discards examples that appear unusual but are likely valuable. The model throws out both tokens from the phrase `Cosmic Microwave', for example, although they are correctly labelled as negative. 

\section*{Appendix C. Implementation Details}
We now describe how to train the robust model using \texttt{glmnet}, starting with the case where the $\theta$ parameters are not penalized. Equation (3) from Section 3.2 shows our reparametrized training objective. It can equivalently be written as 
$$l(\theta') = \sum_{i=1}^n y_ig(\theta'^Tx'_i) + (1 - y_i)\left(1 - g(\theta'^Tx'_i)\right) -  \lambda \sum_{j=0}^{m+n} p_j \abs{\theta'^{(j)}}$$
where $p = (0, \ldots, 0, 1, \ldots 1)$ is a vector of penalty factors, commonly used to allow differential shrinkage. The following code snippet trains such a model for a full path of $\lambda$:
\newline

\begin{verbatim}
robust.train.data = cbind(train.data, diag(N))
penalties = append(rep(0, times=P), rep(1, times=N))
robust.fit = glmnet(robust.train.data, as.factor(train.labels),
    family=`binomial', penalty.factor=penalties, standardize=FALSE)
\end{verbatim}
\ 

When $\theta$ is $L_1$-regularized as well, we instead adopt the following trick. Factoring out $\kappa$ in equation (2) gives us
$$l(\theta, \gamma) = \sum_{i=1}^n y_ig(\theta^Tx_i + \gamma_i) + (1 - y_i)\left(1 - g(\theta^Tx_i + \gamma_i)\right) - \kappa \left( \sum_{j=1}^m \abs{\theta_j} + \frac{\lambda}{\kappa} \sum_{i=1}^n \abs{\gamma_i} \right)$$
Letting $X' = [X | \frac{\kappa}{\lambda}I_n]$ and $\theta' = (\theta_0, \ldots, \theta_m, \frac{\lambda}{\kappa}\gamma_1, \ldots, \frac{\lambda}{\kappa}\gamma_n)$, we can now train the model as usual. If desired, it is simple recover the correct values for $\gamma$. These commands train a regularized model for fixed $\kappa$ and $\lambda$:
\newline

\begin{verbatim}
relative.penalty = lambda / kappa
robust.train.data.local = cbind(train.data, diag(N)/relative.penalty)
robust.fit = glmnet(robust.train.data.local, as.factor(train.labels),
    lambda=kappa, family=`binomial', standardize=FALSE)
\end{verbatim}
\ 

It may seem that one could also use the strategy of supplying a vector of penalty factors, but \texttt{glmnet} internally rescales these factors to sum to $n$. Moreover, the provided technique can be used with practically any software for $L_1$ regularization.

\newpage
\bibliographystyle{plainnat}

\end{document}